\documentclass{article}

\usepackage{arxiv}
\usepackage{float}
\usepackage[utf8]{inputenc} % allow utf-8 input
\usepackage[T1]{fontenc}    % use 8-bit T1 fonts
\usepackage[toc,page]{appendix}      % hyperlinks
\usepackage{url}            % simple URL typesetting
\usepackage{booktabs}       % professional-quality tables
\usepackage{amsfonts}       % blackboard math symbols
\usepackage{nicefrac}       % compact symbols for 1/2, etc.
\usepackage{microtype}      % microtypography
\usepackage{lipsum}		% Can be removed after putting your text content
\usepackage{graphicx}
\usepackage{doi}
\usepackage{csvsimple}
\usepackage{subfig}
\usepackage{amsmath}
\usepackage[numbers]{natbib}
\usepackage{hyperref} 
\usepackage[table]{xcolor}

\title{Hybrid BYOL-ViT: Efficient approach to deal with small Datasets}

\author{Safwen NAIMI$^{1,2}$ \And Rien VAN LEEUWEN \And Wided SOUIDENE$^1$ \And Slim BEN SAOUD$^{1,2}$}
\date{%
    $^1$Tunisia Polytechnic School, University of Carthage, Tunis, Tunisia\\[2ex]%
    $^2$National Institute of Applied Science and Technology, University of Carthage, Tunis, Tunisia\\[2ex]%
    \textcolor{white}{\today}
}

\hypersetup{
pdftitle={SSL and ViT},
pdfsubject={q-bio.NC, q-bio.QM},
pdfauthor={David S.~Hippocampus, Elias D.~Striatum},
pdfkeywords={First keyword, Second keyword, More},
}

\begin{document}
\maketitle

\begin{abstract}
Supervised learning can learn large representational spaces, which are crucial for handling difficult learning tasks. However, due to the design of the model, classical image classification approaches struggle to generalize to new problems and new situations when dealing with small datasets. 
\newline
In fact, supervised learning can lose the location of image features which leads to supervision collapse in very deep architectures. In this paper, we investigate how self-supervision with strong and sufficient augmentation of unlabeled data can train effectively the first layers of a neural network even better than supervised learning, with no need for millions of labeled data. The main goal is to disconnect pixel data from annotation by getting generic task-agnostic low-level features.  Furthermore, we look into Vision Transformers and show that the low-level features derived from a self-supervised architecture can improve the robustness and the overall performance of this emergent architecture. We evaluated our method on one of the smallest open-source datasets STL-10 and we obtained a significant boost of performance from 41.66\% to 83.25\% when inputting low-level features from a self-supervised learning architecture to the ViT instead of the raw images.
\end{abstract}

% keywords can be removed
%\keywords{First keyword \and Second %keyword \and More}

\section{Introduction}

Computing effective image representations is a major challenge nowadays in computer vision, and making the best use of these representations is a long standing topic. One major approach is the supervised learning. This technique requires training on a huge amount of labeled data, which is not always available or accessible in many domains especially in the medical field. To overcome this problem of insufficient labeled training data, some studies \cite{zhuang2020comprehensive,farahani2021concise,shin2016deep} propose the transfer learning technique as a solution. It consists of pre-training some models on large and well established datasets like ImageNet \cite{russakovsky2015imagenet} and JFT-300M \cite{sun2017revisiting}, and then transfer the knowledge to another model. The major problem with this technique is that it only works if the initial and target tasks of both models are similar enough. 
\newline
Moreover, collecting and labeling data for a specific task is time and money consuming process. 
Thus, some researchers are designing architectures that can learn useful representations from a set of unlabeled data \cite{gidaris2018unsupervised,chen2020improved,goodfellow2014generative}. These architectures are based on self-supervised learning where the results are obtained by models that analyze unlabeled data and categorize information independently without any human input.

In this paper, we investigate a new approach for image representation using self-supervised learning to reduce the demand for large labeled data. More specifically, we focus our work on disconnecting pixel data from annotation. We will process by using self-supervised learning techniques to only extract generic task-independent low-level features from unlabeled data. We believe that these are the features capable of boosting the learning on many downstream tasks without dependency to a specific task.

We first explore whether a self-supervised learning technique (BYOL)\cite{grill2020bootstrap} could train effectively the first layers of a convolutional neural network better than fully supervised learning. Our focus is on the first layers since they are responsible of extracting the low-level features from an image. These low-level features are not biased by annotations and labels compared to mid-level features and high-level features. At this level, the features are not related to a specific task and they are highly generic. Secondly, we look over emergent Vision Transformers (ViTs) \cite{dosovitskiy2021image}, which have been proven to achieve comparable or even superior performance on image classification tasks but only when pre-trained on large labeled JFT-300M dataset \cite{sun2017revisiting}. This is likely due to the fact that ViT needs to learn some local proprieties of the visual data using more samples than a Convolutional Network. In order to solve this high dependency to large datasets, we aim to take advantage from ViT's self attention mechanism, but by deploying it together with the power of self-supervised algorithms on extracting generic low-level details. 

Therefore, we implement in the second part the ViT with inputting the low-level features extracted from a fully self-supervised ConvNet.
 We prove in experiments that this approach can effectively attend to important, task-independent features of the image that ViTs struggle to generate from a small labeled data. The overall performance of this approach is comparable to the state of the art without the need for a large amount of labeled data.

The rest of the paper is organized as follows. In section \ref{related}, we propose an overview of related works. In section \ref{secc}, we detail our proposed approach for self-supervised low-level feature extraction. In section \ref{others}, we test our approach with standard ViTs and second-generation ViTs (Compact Transformers \cite{hassani2021escaping}). We conclude in Section \ref{conclusion}. 

In summary, this paper has two goals, first, use the BYOL to extract low-level features instead of a traditional supervised learning technique, second, enhance the performance of a ViT by feeding it with the extracted features from BYOL instead of the row image. The contributions of this paper are the following:
\begin{itemize}
    \item \textit{(1)} We show that learning low-level features from unlabeled images using self-supervised learning can consistently improve the accuracy of a supervised fine-tuning task for both ConvNets and ViTs.
    \item \textit{(2)} We introduce a modification to BYOL's Multi-Layer Perceptron (MLP) architecture that consistently improve the performance and the training time with small datasets.
\item \textit{(3)} We show in experiments that for ViTs, the choice of patch size and data augmentation is important. 

\item \textit{(4)} We show in detail how to overcome the poor performance of first and second generation ViTs with small datasets by relying on a Hybrid BYOL-ViT architecture and how this architecture can outperform the pre-training of ViTs on huge labeled datasets.

\end{itemize}

\section{Related Works}
\label{related}

Our paper relates to four broad areas of research:
 (a) Convolutional Neural Network, (b) Vision Transformer, (c) Self-Supervised Learning and (d) Low-level feature extraction. The following section deals with related work in each one of these fields.

\subsection*{Convolutional Neural Network}
In the realm of deep learning, a convolutional neural network (CNN) is a sophisticated machine-learning approach \cite{gu2017recent}. CNNs use kernels to aggregate very local information in each layer, which is then passed to the next layer that aggregates again local information but with a large field of view. So it starts to look very locally and their receptive field become more global in each layer and after quite some training time. Mainly, CNNs are hard-coded to attend only locally in the lower layers where all the neighboring pixels are related to each other and all parts of the image are processed in the same way regarding of their absolute position. However, CNNs come with some limitations in long range interdependence that self attention mechanisms provide \cite{raghu2021vision}. Meaning that even when CNNs have this ability to attend locally and capture local relationships, they still lack the capacity to track global information when working with small datasets.
%\lipsum[4] See Section \ref{sec:headings}.
\subsection*{Vision Transformer}
ViTs look at the image by taking the input image and splitting it into patches of fixed size \cite{dosovitskiy2021image}. These patches are then flattened and all the work is done in the transformer encoder that contains a MHA mechanism and a MLP with residual connections and layer normalization in between. However, a huge amount of labeled data is needed for training such a network to outperform CNN. In fact, ViTs achieved state-of-the-art on ImageNet classification by directly applying Transformers with global self-attention to full-sized images. A recent study \cite{chen2021transunet} proved that inputting the original image to the ViT leads to limited localization abilities which is due to the limited low-level details. Diving deeply into the details of this architecture shows that the lack of inductive biases present in the CNN architecture is one of the main reasons behind this limited behaviour \cite{raghu2021vision}. It has been proven that when working with a relatively small dataset, the heads of the first self attention layers within the transformer encoder can only attend to large distances and loose the ability to attend to small distances. As a result, the network can only attend to global information of the image in the shallow layers and also the deep layers, causing a poor performance with small datasets. That can explain the reason behind pretraining on large JFT-300M datasets, ViTs are in need for a large number of images to be able to capture local information and track relationship between all patches.
\subsection*{Self-Supervised Learning}
Self-supervised learning is a means for training computers to do tasks without humans providing labeled data. It is a subset of unsupervised learning where outputs are derived by machines that label, categorize and analyze information on their own then draw conclusions based on connections and correlations \cite{site1,henaff2020dataefficient,he2020momentum,chen2020simple}. 
The basic concept of self-supervision relies on encoding an image successfully. A computer capable of self-supervision must know the different parts of any object so it can recognize it from any angle.
A large body of work of self-supervised learning focuses on contrastive learning of visual representations \cite{misra2019selfsupervised,li2021prototypical,bachman2014learning,noroozi2017unsupervised,ledig2017photorealistic,laine2017temporal,Lee_pseudo-label:the,he2020momentum,caron2019deep} and clustering methods \cite{bautista2016cliquecnn,ji2019invariant,asano2020selflabelling,caron2021unsupervised,vangansbeke2020scan}. The goal of contrastive representation learning is to create an embedding space where comparable sample pairs are close together and dissimilar sample pairs are far away. Contrastive learning is one of the most potent ways in self-supervised learning when working with unsupervised data, it aims to learn unsupervised features without differentiating between images. One of the recent methods is SimCLR \cite{chen2020simple}, which is trained by lowering the gap between representations of different augmented views of the same image and increasing the distance between representations of enhanced views from different images, causing positive samples to attract each other and negative samples to repel each other. BYOL \cite{grill2020bootstrap} follows the same contrastive objective and achieves state-of-the art results without using negative pairs, by relying on two neural networks that interact and learn from each other. 

\subsection*{Low-level feature extraction}
Prior to the development of the convolutional neural network, SIFT \cite{Lowe:2004:DIF:993451.996342}, SURF \cite{Meghri2014,sameh2012video,megrhi2013spatio} and HOG \cite{1467360} have been widely used. They are a way for computing regions of interest or key points using Difference of Gaussian (DoG) and then compute the gradients in the local neighborhood and take it into account in order to compute a descriptor vector that can be used to describe or identify the key points. Major advantages of these techniques are robustness to geometrical transformations, occlusion, clutter and extensibility to a wide range of different feature types. Thanks to their invariance to image scale and rotation, those methods have been widely used and extended in \cite{brock2019large,goodfellow2014generative,hessel2017rainbow, vanhasselt2018deep,lillicrap2019continuous}. They have already shown that low-level features can be handcrafted.
According to recent studies \cite{oyallon2017scaling,asano2020critical}, replacing low-level layers in a convolutional neural network with handcrafted features can still reduce the model's overall performance. 
Modern vision systems use supervision to learn these low-level features, which leads to supervision collapse, and loss of any information that isn't required for performing the training job. Unfortunately, such information may be required for transfer to new tasks or domains \cite{doersch2021crosstransformers}. In that case the network, only, picks up on image patterns that tightly group images of the same class in the same feature space and, thus, causing the loss of any information that might be needed for generalization.

%In the following section in this paper is to show that self-supervision can extract meaningful and generic low-level features that can serve as a powerful basis for further downstream tasks. We succeed to completely disconnect the learning of these low-level features from annotations by relying on a bunch of easy to collect unlabeled data. 

\section{Boosting Learning using self-supervision of low-level features}
\label{secc}
In this section, our main goal is to compare the low-level features trained on a supervised way with the low-level features trained using self-supervision. We want to see how far we can go when relying on self-supervised learning. In a pretraining step, we train a network on unlabeled data using self-supervised learning. Then, we freeze the weights of a specific convolution layer from the SSL model and continue the training of the remaining weights using a supervised model. We pick BYOL \cite{grill2020bootstrap} as a self-supervision technique with ResNet18 \cite{he2015deep} as a backbone for the pretraining step. The selection of dataset, network architecture details, data augmentation methods and hyper parameters settings will be discussed in the following subsections.
\subsection{Dataset}
\begin{figure}[!ht]
\begin{center}
\subfloat{\includegraphics[width=0.3\textwidth]{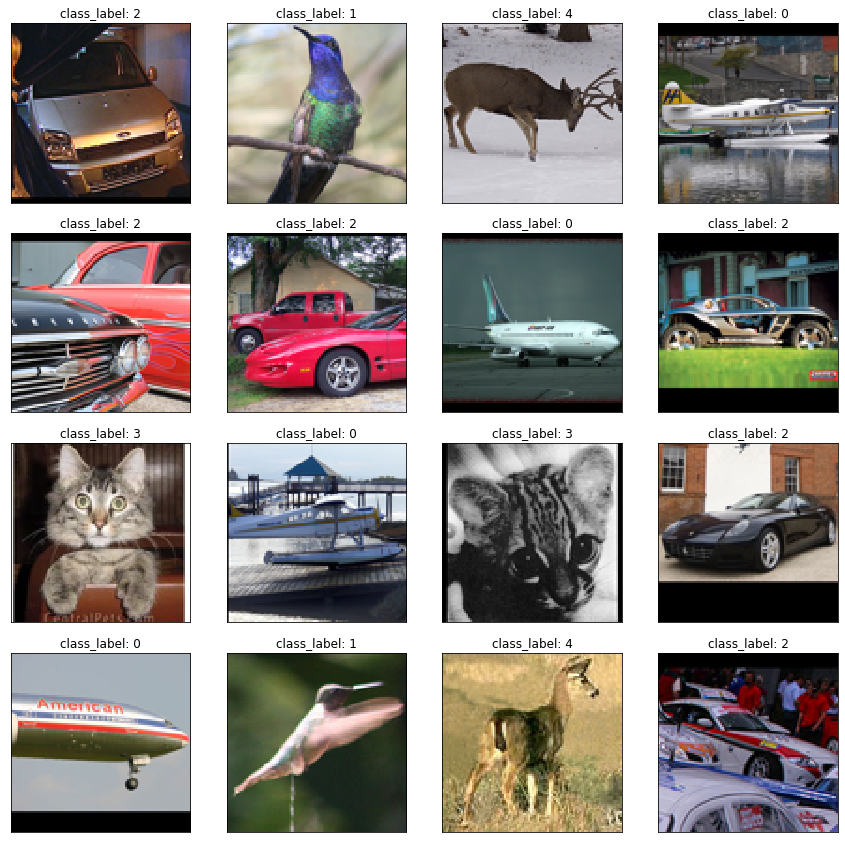}\label{fig:stll}}\quad 
\subfloat{\includegraphics[width=0.31\textwidth]{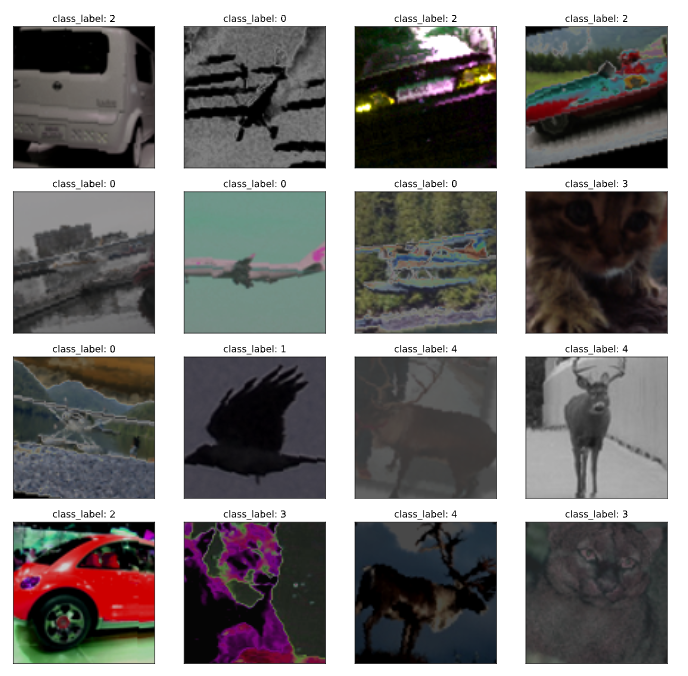}\label{stl_aug}}
\caption{(a) Images from five classes of STL-10 dataset (b) Some training-data augmented STL-10 object images}
\label{fig:all}
\end{center}
\end{figure}

We perform transfer via fine-tuning on the STL-10 dataset which is an image recognition dataset for developing unsupervised feature learning, deep learning and self-taught learning algorithms \cite{site2}. The dataset contains 10 classes with 96×96 pixels colored images. 
For the upcoming experiments, we opted for an extremely tiny dataset, the number of classes is five as shown in Figure \ref{fig:all}. The training data contains 2.2k images belonging to only five classes of STL-10.

\subsection{Data Augmentation}
Data augmentation plays an important role in making self-supervised learning performant \cite{grill2020bootstrap,chen2020simple,he2020momentum}. Creating different transformations of the same image using data augmentation techniques is helpful to provide more possible variations, to minimize over-fitting and to guarantee the generalization of our models. The data augmentation we need for self-supervised learning differs from the data augmentation used for supervised learning. Using excessive and intensive data augmentation for BYOL makes the task harder, so that the network learns many different kinds of features. These transformations need to be chosen carefully because the techniques implemented in data augmentation rely heavily on the dataset.
During this first part, we decided to only implement the same data augmentation mentioned in the original paper \cite{grill2020bootstrap}. We used Color Jitter, Center Cropping, Random Rotation, Random Flipping, Random Grayscale, Gaussian Blur and Solarizing as shown in Figure \ref{fig:all} (See the first row of Table \ref{tab:tab22} for parameter details). Color Jitter and Image Cropping are the most relevant techniques as proved in many previous works \cite{chen2020simple,grill2020bootstrap}. Later in this paper, we will investigate more in depth study about data augmentation. 

\subsection{Experimental results}
\begin{figure}[!ht]
\centering
\includegraphics[width=1.0\textwidth]{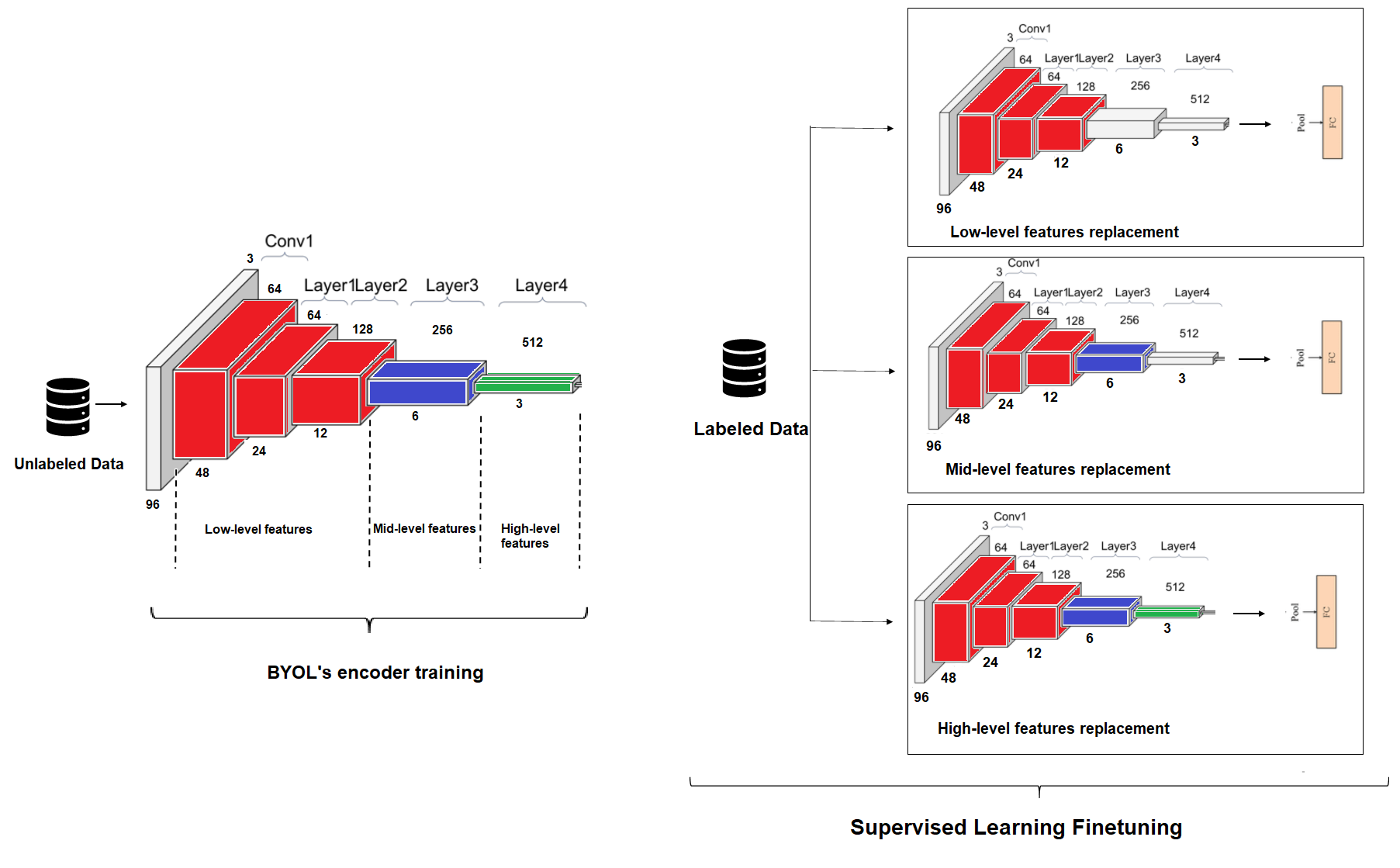}
\caption{Overview of the approach}
\label{fig:arch}
\end{figure}

To test the robustness and effectiveness of the features extracted from a self-supervised training, we opted for this procedure: first, we started by training BYOL with ResNet18 as a backbone on different sets of unlabeled data\cite{site2}. 
\newline
After training BYOL’s model, we freeze a block of layers each time from the encoder architecture (ResNet18) and we fine tune a supervised model using the obtained BYOL's knowledge.
\newline
Our goal was to show that BYOL is able to boost the learning of the supervised model after transfer of knowledge even when trained on a relatively small dataset. We wanted to measure the quality of shallow and deep features on a per layer basis and demonstrate how far we can go with unlabeled data.
\newline
Our proposed approach is simplified in Figure \ref{fig:arch}.

In Table \ref{tab:sslssl}, we report the accuracy after retraining ResNet18 on a fully supervised manner with n convolutional layers left frozen from BYOL's backbone: n $\in$ [1 $\dots$ 4].
\newline
Comparing the results of Table \ref{tab:addlabel} and Table \ref{tab:sslssl}, we observe that using BYOL knowledge successfully boosted the learning of the supervised model in two cases: replacing the low-level features and the mid-level features. 
We can note that the accuracy depends on the number of images used to train both methods: SL and SSL, and that self-supervised learning always boosts the learning of the supervised model when using the features extracted from the shallow layers of BYOL. 
As shown in Table \ref{tab:sslssl}, training BYOL on the 100k unlabeled images provided with STL-10, then using the entire block before layer3 as a low-level feature extractor by freezing it, and fine-tune the rest of the supervised model boosts the accuracy from 59.8\% to 71.96\%. Approximately 12\% accuracy boost.
\newline
\newline
\newline
% Table generated by Excel2LaTeX from sheet 'Sheet3'
\begin{table}[ht]
  \centering
  \caption{Supervised learning from scratch results on different portions of the training data}
    \begin{tabular}{|cc|r|r|r|r|r|r|}
    \toprule
    \multicolumn{2}{|p{12.27em}|}{Amount of Labeled Data\newline{}used to train SL from scratch} & \multicolumn{1}{p{4.045em}|}{100\% of \newline{}training Data} & \multicolumn{1}{p{4.045em}|}{75\% of \newline{}training Data} & \multicolumn{1}{p{4.045em}|}{50\% of \newline{}training Data} & \multicolumn{1}{p{4.045em}|}{25\% of \newline{}training Data} & \multicolumn{1}{p{4.045em}|}{5\% of \newline{}training Data} & \multicolumn{1}{p{4.045em}|}{1\% of \newline{}training Data} \\
    \midrule
    \multicolumn{2}{|c|}{Top-1 Accuracy} & 59.80\% & 55.67\% & 56.45\% & 46.32\% & 13.69\% & 18.33\% \\
    \bottomrule
    \end{tabular}%
  \label{tab:addlabel}%
\end{table}%
\newline
\newline
\begin{table}[H]
  \centering
  \caption{Top-1 Accuracy after features replacement from BYOL's knowledge}
    \begin{tabular}{|l|r|r|r|r|r|r|r|}
    \toprule
    \multicolumn{1}{|p{13.455em}|}{Amount of Unlabeled\newline{}Data used to train BYOL} & \multicolumn{1}{p{4.045em}|}{100k} & \multicolumn{1}{p{4.045em}|}{100\% of \newline{}training Data} & \multicolumn{1}{p{4.045em}|}{75\% of \newline{}training Data} & \multicolumn{1}{p{4.045em}|}{50\% of \newline{}training Data} & \multicolumn{1}{p{4.045em}|}{25\% of \newline{}training Data} & \multicolumn{1}{p{4.045em}|}{5\% of \newline{}training Data} & \multicolumn{1}{p{4.045em}|}{1\% of \newline{}training Data} \\
    \midrule
    \multicolumn{1}{|p{13.455em}|}{Amount of Labeled Data\newline{}used to train SL finetuning} & \multicolumn{1}{p{4.045em}|}{2.2k\newline{}} & \multicolumn{1}{p{4.045em}|}{100\% of \newline{}training Data} & \multicolumn{1}{p{4.045em}|}{75\% of \newline{}training Data} & \multicolumn{1}{p{4.045em}|}{50\% of \newline{}training Data} & \multicolumn{1}{p{4.045em}|}{25\% of \newline{}training Data} & \multicolumn{1}{p{4.045em}|}{5\% of \newline{}training Data} & \multicolumn{1}{p{4.045em}|}{1\% of \newline{}training Data} \\
    \midrule
    High-level features replacement & 62.93\% & 52.15\% & 53.75\% & 54.05\% & 44.33\% & 34.53\% & 32.50\% \\
    \midrule
    Mid-level features replacement & 69.54\% & 66.20\% & 65.64\% & 64.53\% & 51.29\% & 37.90\% & 25.33\% \\
    \midrule
    Low-level features replacement & 71.96\% & 66.25\% & 65.11\% & 61.76\% & 51.44\% & 27.53\% & 16.33\% \\
    \bottomrule
    \end{tabular}%
  \label{tab:sslssl}%
\end{table}%
.
\newline
\newline
\newline

\subsection{BYOL modification}
\begin{figure}[!ht]
\begin{center}
\subfloat[]{\includegraphics[width=0.3\textwidth]{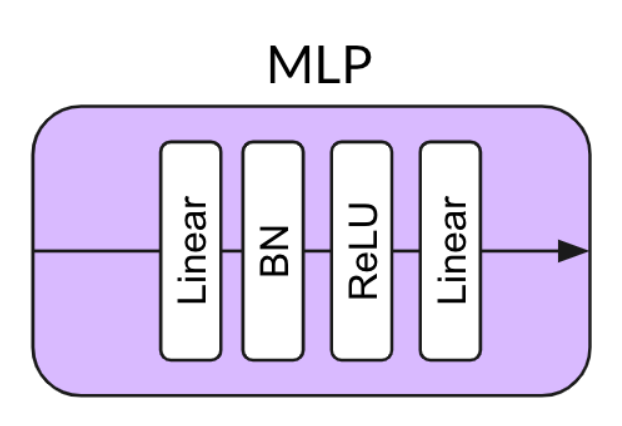}\label{fig:proj}}\quad 
\subfloat[]{\includegraphics[width=0.3\textwidth]{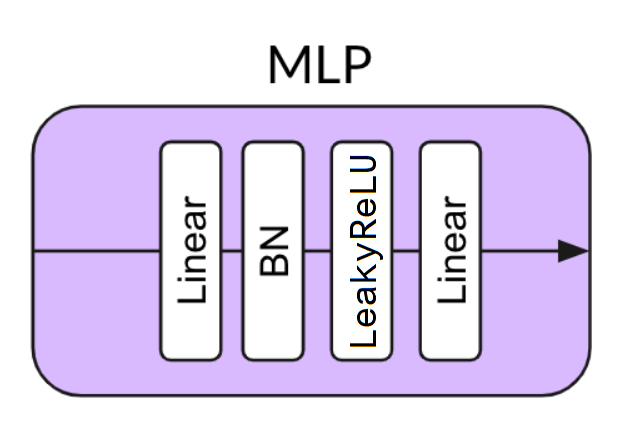}\label{fig:mlp}}
\caption{(a) Original BYOL's MLP (b) Our modified BYOL's MLP}
\label{fig:our}
\end{center}
\end{figure}

In the original architecture of BYOL, authors used an MLP composed of a Linear layer followed by a BN, ReLU layer and a final Linear layer. 
Some researches \cite{xu2015empirical} proved that sparsity is the key for good performance in ReLU. 
Moreover, on small scale datasets like STL-10, using deterministic negative slope or learning it are both prone to overfitting. The downside for giving zero to all negative values is a problem called "dying ReLU" \cite{Lu_2020}.
A ReLU neuron is dead if it's stuck in the negative side and always outputs 0. Over the time, the model might end up with a large part doing nothing. Our modification suggests that incorporating a non-zero slope for negative part in the MLP architecture of BYOL as shown in Figure \ref{fig:our} could consistently improve the results.
The leaky rectified linear function LeakyReLU has a small slope for negative values, instead of altogether zero.
\newline
The results of replacing ReLU with LeakyReLU in the MLP architecture of BYOL are reported in Table \ref{tab:tab3}. 
\newline
During these experiments, we trained BYOL on 2.2k unlabeled data and on 100k unlabeled data for 100 epochs, then we froze layer2 and fine-tuned a supervised model trained on 2.2k labeled data.
\newline
% Table generated by Excel2LaTeX from sheet 'Sheet4'
\begin{table}[htbp]
  \centering
  \caption{Comparison between the original MLP architecture and our modified architecture when trained on different portions of unlabeled data. We reported Top-1 Accuracy after fine-tuning the supervised model.}
    \begin{tabular}{|l|c|c|c|}
    \toprule
    \textbf{Model} & \textbf{\#images} & \textbf{Top-1 accuracy}\\
    \midrule
    BYOL (ReLU) & 2200  & 66.25\%  \\
    \midrule
    BYOL (LeakyReLU) & 2200  & 67.31\% \textcolor{red}{(+1.06\%)}  \\
    \midrule
    \midrule
    BYOL (ReLU) & 100k  & 69.54\%  \\
    \midrule
    BYOL (LeakyReLU) & 100k  & 72.86\% \textcolor{red}{(+3.32\%)}  \\
    \bottomrule
    \end{tabular}%
  \label{tab:tab3}%
\end{table}%
\newline
From the results of Table \ref{tab:tab3}, we can say that LeakyReLU instead of ReLU in the MLP architecture has two benefits for our training despite fixing the dying ReLU problem: We ended up with a consistent boost in performance.
More in-depth studies and experiments about the efficiency of replacing ReLU with LeakyReLU can be found in Appendix \ref{appendix:d}. In the remainder of this paper, we will keep using LeakyReLU instead of ReLU in the MLP architecture of BYOL.

\section{Hybrid BYOL-ViT architecture}
\label{others}
In the previous section, we showed that the approach of generating low-level features from a set of unlabeled data using BYOL is capable of boosting the overall performance of ConvNets. In this section, we question whether this is also beneficial for the new emerging ViTs. We first look into the standard ViT architecture and why this approach can be a solution. Then we comparatively analyze the efficiency and
accuracy under different conditions.
\subsection{ Implementation results}
\begin{figure}[!ht]
\centering
\includegraphics[width=1.0\textwidth]{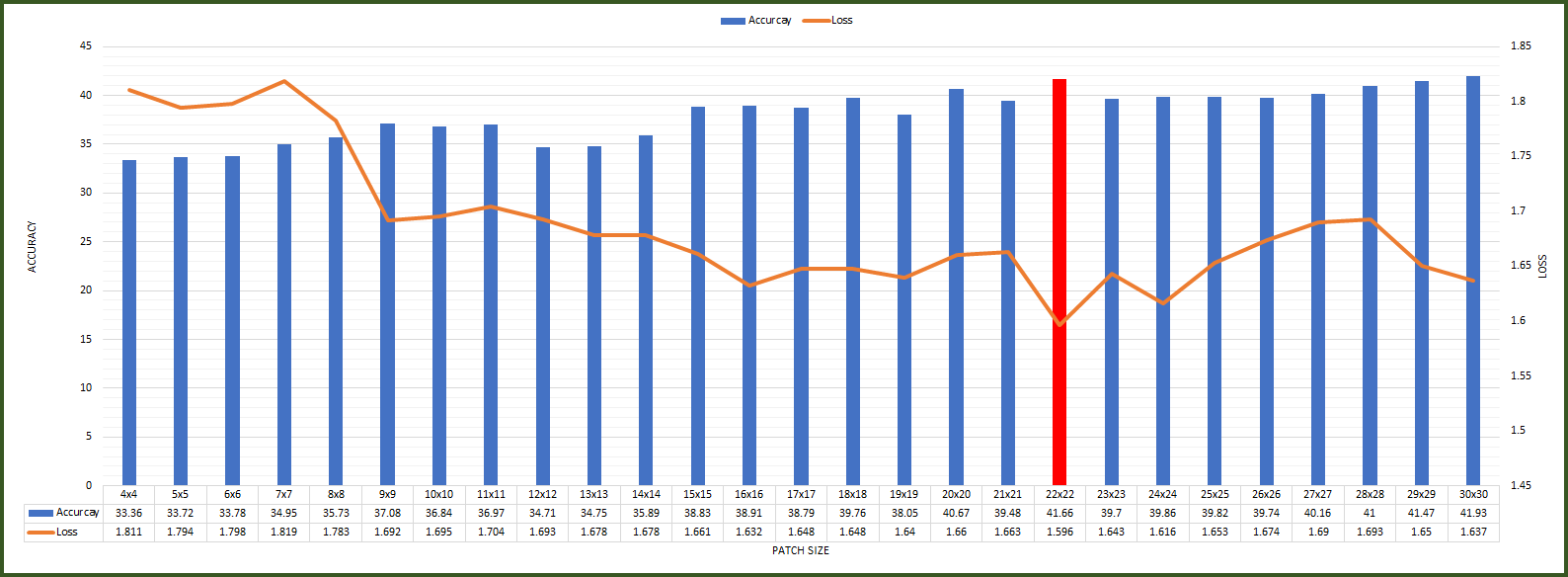}
\caption{Results of training a Large ViT from scratch on STL-10 dataset with every possible patch size}
\label{fig:results}
\end{figure}

\begin{figure}[!ht]
\centering
\includegraphics[width=1.0\textwidth]{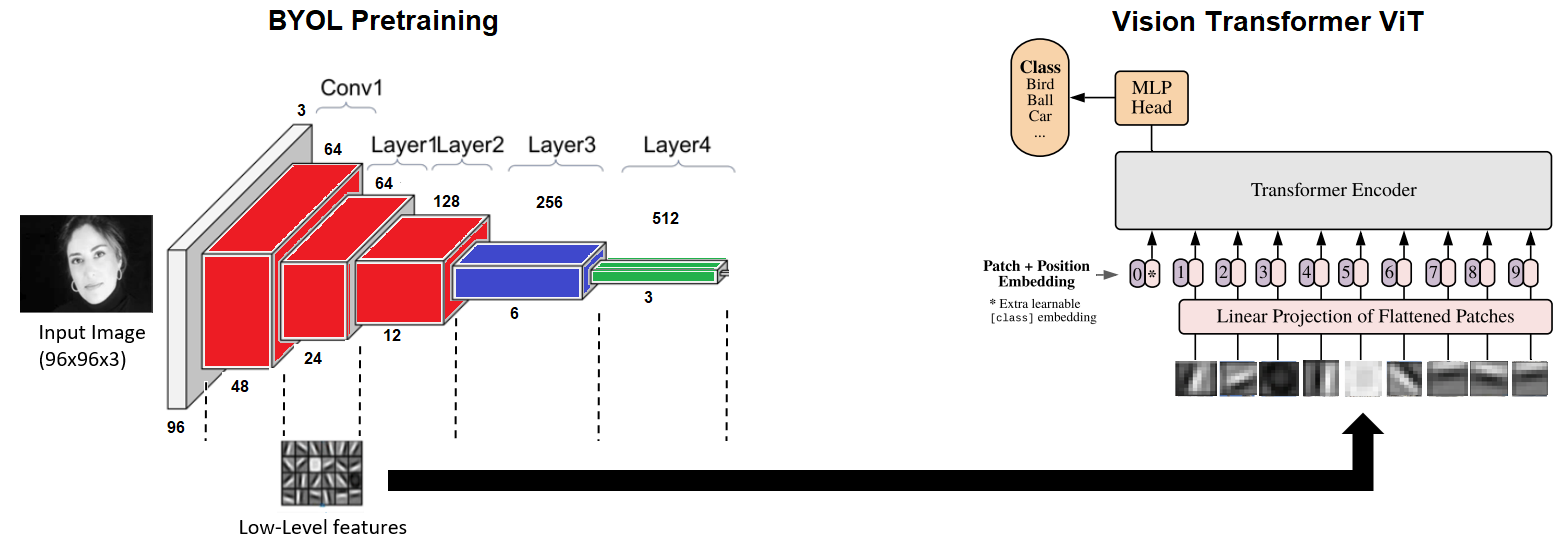}
\caption{Overview of the proposed approach when inputting the features extracted after BYOL's layer1 to a standard ViT}
\label{fig:archi}
\end{figure}

We started by evaluating a standard ViT on STL-10 dataset for 600 epochs, with a batch size of 128. The optimizer is Adam with a learning rate of 0.0001 and a weight decay of 0.05. For the data augmentation, we opted for Random Crop, Random Horizontal Flip and Gaussian Blur. In the original paper of ViT \cite{dosovitskiy2021image}, they had this assumption that the best patch size is 16x16 since it results in a larger effective sequence length. 
We wanted to know if this applies to STL-10 dataset as well.
The results of our experiments are shown in Figure \ref{fig:results}. We observe that the best accuracy when training the standard ViT on STL-10 dataset from scratch and without pretraining is obtained with patch size of 22x22: Top-1 Accuracy of 41.66\% and a loss of 1.596. We hypothesize that the mediocre performance on STL-10 dataset is due to the limited low-level details since we are dealing with a small dataset. Recent work \cite{raghu2021vision} proved that the ViT is really data hungry. Training on a relatively small dataset like STL-10 leads to lower layers not learning to attend more locally, resulting in an overall ViT that can only attend to global information of the image and lose the ability to track close patches. Our thinking is that low-level features containing local information are relevant for strong performance and that the inability of ViT to extract them from a tiny image distribution is the main reason behind that poor performance.
\newline
To improve the emerging ViT models, recent researches \cite{dosovitskiy2021image,chen2021transunet} proved that when combining CNN with ViT, we can obtain better results. In our work, instead of inputting the feature map from a fully supervised model, we suppose that generating features from a set of images using self-supervised learning leads to a richer feature content as proved in the previous section \ref{secc}.
For that, we opted for the architecture presented in Figure \ref{fig:archi}, we extracted features from different layers of BYOL's backbone. 
These features are going to serve as input to the ViT instead of the real row image. 
We want to test if the features trained on a fully self-supervised manner are capable of boosting the performance of ViTs. Our goal is to overcome the inability of ViT's first layers to track local information by inputting them from a self-supervised model without the need for pretraining this ViT on large JFT-300M.
\newline
\newline
\newline
\newline
To test the efficiency of our hybrid architecture, we trained BYOL on the set of unlabeled data provided with STL-10 dataset: 100k images and using ResNet as a backbone.
We opted for Adam optimizer \cite{kingma2017adam} with a batch size of 16 (See Appendix \ref{appendix:f}), a learning rate of 10$^{-4}$ and a weigh decay of 5.10$^{-2}$ (See Appendix \ref{appendix:e}).
For the data augmentation used to train BYOL, we opted for a modified version compared to the one mentioned in the original paper \cite{grill2020bootstrap}. The results of all experiments comparing data augmentations used to train BYOL are reported in Appendix \ref{appendix:abc}.
After finding the best parameters to train BYOL, we used the features extracted after each layer of the encoder as input to the ViT. We did multiple experiments also to find the adequate data augmentation for the supervised part. Results of these experiments can be found in Appendix \ref{appendix:b}. 
The overall results of training a hybrid BYOL-ViT architecture with the optimal BYOL parameters are collected in Table \ref{tab:compar}. We report Top-1 accuracy and  Loss for each experiment.
\begin{table}[htbp]
  \centering
  \caption{Results of our Hybrid BYOL-ViT architecture with features extracted from different layers of the BYOL's backbone (ResNet50) and with every possible patch size. BYOL trained using data\_aug\_5 for 400epochs. (See Table \ref{appendix:ab})}
    \begin{tabular}{|c|l|r|r|}
    \toprule 
          & \textbf{Patch Size} & \multicolumn{1}{l|}{\textbf{Top-1 Accuracy}} & \multicolumn{1}{l|}{\textbf{Loss}} \\
    \midrule
    {\textbf{Features extracted after Layer1}} & 1×1   & 71.92\% & 0.7858 \\
          & 2×2   & 73.17\% & 0.7892 \\
          & 3×3   & 73.17\% & 0.7640 \\
          & 4×4   & 75.33\% & 0.7112 \\
          & 5×5   & 71.33\% & 0.7756 \\
          & 6×6   & 74.17\% & 0.7613 \\
          & 7×7   & 71.83\% & 0.8144 \\
          & 8×8   & 73.67\% & 0.7859 \\
          & 9×9   & 70.17\% & 0.8339 \\
          & 10×10 & 71.08\% & 0.8146 \\
          & 11×11 & 74.25\% & 0.8221 \\
          & 12×12 & 73.75\% & 0.7975  \\
          & 14×14 & 64.33\% & 0.9982  \\
          & 16×16 & 66.08\% & 0.9458 \\
          & 18×18 & 68.58\% & 0.8949 \\
          & 24×24 & 72.83\% & 0.8324 \\
    \midrule
    {\textbf{Features extracted after Layer2}} & 1×1   & 83.25\% & 0.4979 \\
          & 2×2   & 81.42\% & 0.5285 \\
          & 3×3   & 80.17\% & 0.5729 \\
          & 4×4   & 81.25\% & 0.5633 \\
          & 5×5   & 78.5\% & 0.628 \\
          & 6×6   & 80.58\%  & 0.6079 \\
          & 7×7   & 74.42\% & 0.7421 \\
          & 8×8   & 78.83\% & 0.647 \\
          & 9×9   & 78.33\% & 0.6363 \\
          & 10×10 & 80.33\% & 0.6185 \\
          & 11×11 & 77.5\% & 0.6305 \\
          & 12×12 & 78.17\% & 0.5995 \\
    \midrule
    {
    \textbf{Features extracted after Layer3}} & 1×1   & 83.42\% & 0.4901 \\
          & 2×2   & 83.58\% & 0.4749 \\
          & 3×3   & 83.17\% & 0.4973 \\
          & 4×4   & 82.08\% & 0.5503 \\
          & 5×5   & 80.75\% & 0.5345 \\
          & 6×6   & 82.75\% & 0.4992 \\
    \midrule
    {\textbf{Features extracted after Layer4}} &  1x1   & 81.42\% & 0.558 \\
          & 2x2   & 79.58\% & 0.5916 \\
          & 3x3   & 78.92\% & 0.5796 \\
    \bottomrule
    \end{tabular}%
  \label{tab:compar}%
\end{table}%

The results have exposed the potential of BYOL to extract robust low-level features. We were able to improve the accuracy of a standard ViT from 41.66\% to 83.25\% when inputting the low-level features extracted from Layer2 of BYOL’s backbone with patch size of 1x1 instead of the real row image. We found also that providing the mid-level features and the high-level features extracted from Layer3 and Layer4 respectively leads to an overfitting of the overall architecture. Our contribution was providing the much-needed low-level features that standard ViT struggles to generate when dealing with a small dataset.
Another finding from these experiments is that patch size in ViT decides the length of the sequence. Lower patch size leads to higher information exchange during the self-attention mechanism. This is verified by the better results using lower patch-sizes. 
%%We have the assumption that when working with a hybrid BYOL-ViT architecture, lower patch sizes always works better.
%%As we discussed in the previous section \ref{secc}, data augmentation is an important ingredient for learning robust and generic representations, this is the key for making vision transformers work well with small datasets since we were able to overcome the limitations of ViT for capturing local information by inputting them from BYOL knowledge, that's a sort of distillation between a CNN architecture and a ViT architecture. 
%%We still believe that deeper research on more adapted or learned data augmentation will bring further gains.
\subsection{Performance Comparison}
In order to study the effectiveness of our modified BYOL models, we downloaded a checkpoint of BYOL model with ResNet50 as a  backbone pretrained on ImageNet ILSVRC-2012 dataset \cite{russakovsky2015imagenet} which is 13 times bigger compared to the set of unlabeled data provided with STL-10 dataset (100k images). We did a hybrid BYOL-ViT training with features extracted from Layer2 of BYOL's backbone and a patch size of 1x1.
\newline
We compared our BYOL models pretrained only on 100k images with the one downloaded from the original work \cite{grill2020bootstrap} pretrained with ImageNet dataset.
Experiments results are reported in Table \ref{tab:per_comp}.
% Table generated by Excel2LaTeX from sheet 'Recap'

\begin{table}[htbp]
  \centering
  \caption{Top-1 Accuracy and Loss comparison between an original BYOL\cite{grill2020bootstrap} (trained on ImageNet) and our BYOL (trained on 100k images). Features are then extracted from layer2 and fed to the Large ViT \cite{dosovitskiy2021image}.}
    \begin{tabular}{|l|l|l|r|r|}
    \toprule
    \textbf{Method} &
    \multicolumn{1}{p{8em}|}{\textbf{SSL pretraining}\newline{}\textbf{data size}} &
    \multicolumn{1}{p{8em}|}{\textbf{Backbone}\newline{}\textbf{architecture}} & \multicolumn{1}{l|}{\textbf{Top-1 Accuracy}} & \multicolumn{1}{l|}{\textbf{Loss}} \\
    \midrule
    BYOL \cite{grill2020bootstrap} - ViT\_Large & 1.3M images & ResNet50 & 81.75\% & 0.5251 \\
    \midrule
    BYOL (ours) - ViT\_Large & 100k images & ResNet50 & 83.25\%  & 0.4979 \\
    \midrule
    BYOL (ours) - ViT\_Large & 100k images & ResNet18 & 80.33\% & 0.5881 \\
  
    \bottomrule
    \end{tabular}%
  \label{tab:per_comp}%
\end{table}%

Our approach for training BYOL with an adequate data augmentation and a LeakyReLU layer greatly improves the performance of the large ViT on STL-10 dataset, using only 13 times less data in the self-supervised learning pretraining part. We found also that in order to extract generic and robust low-level features from BYOL, we need to train it using small batch sizes, which is contradictory with the original paper \cite{grill2020bootstrap} that suggests using larger batch sizes. A complete study on the influence of batch size in the overall performance of our Hybrid BYOL-ViT architecture can be found in Appendix \ref{appendix:f}.

\subsection{Influence of Model Size}
Recent work \cite{chen2020big} mentioned the effectiveness of big models on supervised learning \cite{sun2017revisiting,mahajan2018exploring}, fine-tuning supervised models on a few examples \cite{kolesnikov2020big} and unsupervised
learning on language \cite{raffel2020exploring,devlin2019bert,brown2020language}. We wanted to see the influence of these big models in capturing generic low-level features from an image. Those features, as demonstrated in the previous sections, can successfully boost the learning of ViTs.
In order to study the effectiveness of big models, we trained BYOL with ResNet18, ResNet50 and WideResNet50 \cite{zagoruyko2017wide} as backbone on 100k unlabeled data with the optimal data augmentation which is Data\_aug\_5 (See Table \ref{appendix:ab}) for multiple epochs. The results are reported in Table \ref{tab:big_mod}.

% Table generated by Excel2LaTeX from sheet 'Res18 VS Res50'
\begin{table}[htbp]
  
  \centering
  \caption{Top-1 Accuracy and Loss of Hybrid BYOL-ViT model with BYOL trained using different backbone architectures for multiple epochs. Hybrid architecture trained on STL-10 dataset.}
    \begin{tabular}{|l|r|r|r|r|r|r|}
    \toprule
          & \multicolumn{2}{c|}{\textbf{ResNet18}} & \multicolumn{2}{c|}{\textbf{ResNet50}} & \multicolumn{2}{c|}{\textbf{WideResNet50}} \\
          
    \midrule
    \multicolumn{1}{|p{5.1em}|}{\textbf{\#Epochs to}\newline{}\textbf{train BYOL}} & \multicolumn{1}{l|}{\textbf{Accuracy}} & \multicolumn{1}{l|}{\textbf{Loss}} & \multicolumn{1}{l|}{\textbf{Accuracy}} & \multicolumn{1}{l|}{\textbf{Loss}} & \multicolumn{1}{l|}{\textbf{Accuracy}} & \multicolumn{1}{l|}{\textbf{Loss}} \\
    \midrule
    100epochs & 76.25\% & 0.678 & 76.92\% & 0.6498 & 82.08\% & 0.5472 \\
    \midrule
    200epochs & 78.83\% & 0.6   & 80.33\% & 0.5833 & 82.67\%      & 0.5004 \\
    \midrule
    300epochs & 80.33\% & 0.5881 & 82.17\% & 0.5291 &  83.67\%     & 0.4949  \\
    \midrule
    400epochs & 79.42\% & 0.5904 & 83.25\% & 0.4979 &  83.67\%     & 0.5066  \\
    \bottomrule
    \end{tabular}%
  \label{tab:big_mod}%
\end{table}%

We can see that increasing the depth of BYOL's backbone from ResNet18 to ResNet50 boosts significantly the learning of the hybrid architecture, we obtained an improvement of 2.92\% from 80.33\% to 83.25\%. Also increasing the width of BYOL's backbone from ResNet50 to WideResNet50 boosts the learning of our hybrid architecture with a little improvement of 0.42\% from 83.25\% to 83.67\%.
Our key takeaway from these experiments is that bigger models,
which could easily overfit when trained on a small set of data using supervised learning, can generalize much better and can be more efficient in extracting task-agnostic low-level features from the first layers (Layer2 in our case) when trained on a self-supervised way. In order to further prove this assumption, we extracted features from Layer2 of a supervised ResNet50 pretrained on ImageNet and we fed them to a Large ViT, we did the same with a ResNet50 used as the backbone of BYOL model pretrained also on ImageNet. See Figure \ref{tab:slssl} for results of this comparison.
\begin{table}[htbp]
  \centering
  \caption{Top-1 Accuracy and Loss comparison of a Hybrid ResNet50-ViT architecture between a ResNet50 pretrained
on a supervised way and a ResNet50 pretrained on a self-supervised way using BYOL}
    \begin{tabular}{|l|l|r|r|}
    \toprule
    \textbf{Model} & \multicolumn{1}{p{10.725em}|}{\textbf{\#images to} \newline{}\textbf{pretrain ResNet50}} & \multicolumn{1}{l|}{\textbf{Top-1 Accuracy}} & \multicolumn{1}{l|}{\textbf{Loss}} \\
    \midrule
    ResNet50 (SL)   - ViT\_Large & 1.3M  & 69.2\%  & 0.86 \\
    \midrule
    ResNet50 (SSL) - ViT\_Large & 1.3M  & 81.8\%  & 0.52 \\
    \bottomrule
    \end{tabular}%
  \label{tab:slssl}%
\end{table}%

Results from Table \ref{tab:slssl} show that ResNet50 is more efficient in extracting low-level information from a set of images when trained on a self-supervised way. Same architecture struggles to extract low-level information when trained on a fully supervised manner because of that dependency to annotation. To give more credits to this assumption, we tried also with multiple big models. Results of comparing supervised learning and self-supervised learning when trained with big models are reported in Table \ref{tab:bigg}. 

% Table generated by Excel2LaTeX from sheet 'Recap'
\begin{table}[htbp]
  \centering
  \caption{Influence of model size comparison between supervised learning and self-supervised learning for our approach. SSL is the backbone of our modified BYOL}
    \begin{tabular}{|c|c|c|c|||c|c|c|}
    \toprule
    \textbf{Method} & \multicolumn{3}{c|||}{\textbf{Hybrid SL-ViT}} & \multicolumn{3}{c|}{\textbf{Hybrid SSL-ViT}} \\
    \midrule
    \textbf{CNN architecture} & ResNet50 & WideResNet50 & WideResNet101 & ResNet50 & WideResNet50 & WideResNet101 \\
    \midrule
    \textbf{\#images} & 1.3M  & 1.3M  & 1.3M  & 100k  & 100k  & 100k \\
    \midrule
    \textbf{Top-1 Accuracy} & 69.2\%  & 69.2\%  & 65.08\%  & 83.67\% & 83.25\% & 83.94\% \\
    \midrule
    \textbf{Loss}  & 0.858   & 0.818  & 0.911 & 0.498 & 0.495 & 0.487 \\
    \bottomrule
    \end{tabular}%
  \label{tab:bigg}%
\end{table}%
Results from Table \ref{tab:bigg} proved that big models, if trained on a supervised way, can overfit easily even with relatively large datasets. Same models trained on a self-supervised way generate good representations when dealing with small unlabeled data of 100k images.This is verified by the accuracy of the Hybrid SSL-ViT model. Our key take away is that for a Hybrid SL-ViT, implementing big models does not lead to better results, and extending the number of images does not help. But, for a Hybrid SSL-ViT, implementing bigger and wider models helps to extract more generic low-level features and leads to a significant boost of performance.

\subsection{Experiments with Compact Transformers}
\begin{figure}[!ht]
\centering
\includegraphics[height=0.135\textwidth]{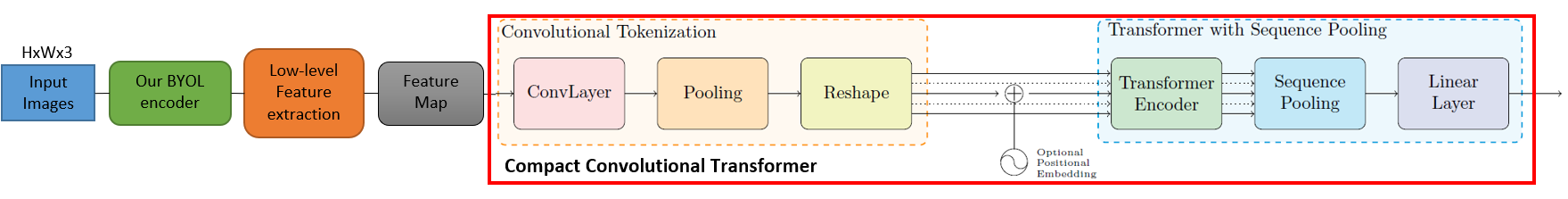}
\caption{Steps to implement our approach with Compact Convolutional Transformers CCT}
\label{fig:cct}
\end{figure}

\begin{figure}[H]
\centering
\includegraphics[height=0.135\textwidth]{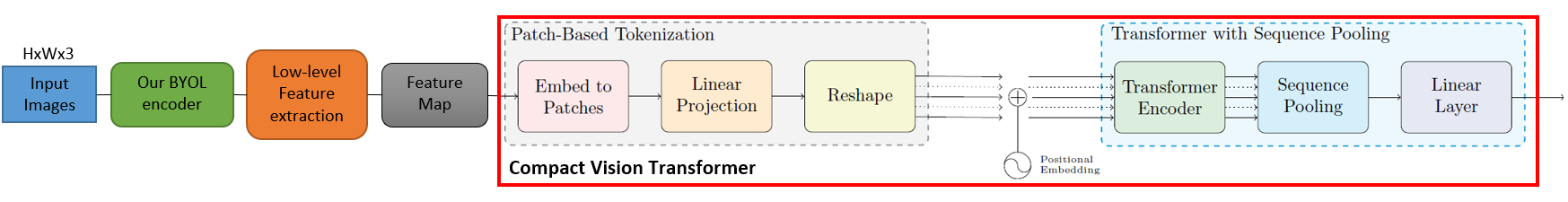}
\caption{Steps to implement our approach with Compact Vision Transformers CVT}
\label{fig:cvt}
\end{figure}

Compact transformers were introduced in the paper of Hassani et al \cite{hassani2021escaping}, the idea was to eliminate the requirement for class token and positional embeddings through a novel sequence pooling strategy that allows the compact ViT to weight the sequential embeddings of the latent space produced by the transformer encoder and better correlate data across the input. That technique allows the ViT to give higher weights to patches that contain more information relevant to the classifier. The paper also introduced a convolutional based paching method called compact convolutional transformer that allow for efficient tokenization
and preserves local spatial relationships. The authors reported competitive results on mid-range datasets like CIFAR-100 \cite{site9} and also on large datasets such as ImageNet \cite{russakovsky2015imagenet}. We wanted to test our approach of inputting low-level features derived from the first layers of our BYOL instead of the real row image to the two versions of compact transformers, CVT and CCT. We will test this approach on tiny STL-10 dataset \cite{site2}. A detailed implementation of our approach with compact convolutional transformers (CCT)  can be found in Figure \ref{fig:cct} and same implementation with compact vision transformers (CVT) in Figure \ref{fig:cvt}.

For this approach, we are inputting the feature map derived from BYOL encoder. Our thinking is that the low-level features learned from a ConvNet trained on a fully self-supervised manner can also boost the final performance of compact transformers.
% Table generated by Excel2LaTeX from sheet 'compact'
\begin{table}[H]
  \centering
  \caption{Top-1 Accuarcy Comparison between a standard Vision Transformer, a CVT and a CCT with and without our approach. All models are trained on STL-10 dataset \cite{site2}}
    \begin{tabular}{|l|c|r|r|}
    \toprule
    \multicolumn{1}{|c|}{Model} & \multicolumn{1}{p{4.775em}|}{\#images to \newline{}train BYOL} & \multicolumn{1}{p{4.045em}|}{Top-1\newline{}Accuracy} & \multicolumn{1}{c|}{\#Params} \\
    \midrule
    \multicolumn{1}{l}{Vision Transformer} & \multicolumn{1}{r}{} & \multicolumn{1}{r}{} & \multicolumn{1}{r}{} \\
    \midrule
    ViT\_Large pretrained with JFT-300M &  ---    & 94.14\% & 306M \\
    ViT\_Large from scratch &  ---    & 41.66\% & 32.75M \\
    BYOL(R18)-ViT\_Large & 100k  & 80.33\% & 32.07M \\
    BYOL(R50)-ViT\_Large & 1.3M  & 83.25\% & 32.27M \\
    \midrule
    \multicolumn{1}{l}{Compact Vision Transformer} & \multicolumn{1}{r}{} & \multicolumn{1}{r}{} & \multicolumn{1}{r}{} \\
    \midrule
    CVT-2/1  & ---      & 74.36\% & 0.28M \\
    BYOL(R18)-CVT2/1 & 100k  & 83.22\% & 0.8M \\
    BYOL(R50)-CVT2/1 & 1.3M  & 87.55\% & 2.37M \\
    \midrule
    CVT-7/4  &  ---     & 73.01\% & 3.85M \\
    BYOL(R18)-CVT7/4 & 100k  & 82.97\% & 4.88M \\
    BYOL(R50)-CVT7/4 & 1.3M  & 87.61\% & 8.03M \\
    \midrule
    \multicolumn{1}{l}{Compact Convolutional Transformer} & \multicolumn{1}{r}{} & \multicolumn{1}{r}{} & \multicolumn{1}{r}{} \\
    \midrule
    CCT6/3x1 & ---      &  ---     & --- \\
    BYOL(R18)-CCT6/3x1 & 100k  & 87.59\% & 3.75M \\
    BYOL(R50)-CCT6/3x1 & 1.3M  & 92.35\% & 5.52M \\
    \midrule
    CCT2/3x1 & ---      & 66.90\%      & 0.48M  \\
    BYOL(R18)-CCT6/3x1 & 100k  & 85.11\% & 0.49M \\
    BYOL(R50)-CCT6/3x1 & 1.3M  & 92.04\% & 1.38M \\
    \bottomrule
    \end{tabular}%
  \label{tab:cvtcct}%
\end{table}%

Despite the fact that compact transformers work well with mid-range datasets like CIFAR-10 \cite{site9} as mentioned in the original paper \cite{hassani2021escaping}, results from Table \ref{tab:cvtcct} show that the performance can be poor when dealing with tiny datasets like STL-10 \cite{site2}.  Similar to the results with standard ViTs, we found that our approach works extremely well with compact transformers also. Results from Table \ref{tab:cvtcct} show that our approach of inputting low-level features derived from a set of unlabeled data can significantly boost the performance when dealing with small datasets. The results are competitive to a ViT Large pretrained on JFT-300M dataset with less resource usage.
\section{Conclusion}
\label{conclusion}
In this paper, we suggest that the best way to learn good representations is by using the unlabeled data to only learn the low-level features and in that case the representations are not directly
tailored to a specific classification task. With this task-independent use of unlabeled data, we found that using the low-level features extracted from the first layers of BYOL's backbone greatly improves accuracy of supervised models.
We further investigate with experiments a new architecture of BYOL’s MLP and we reported a significant boost of performance compared to the basic MLP architecture. Our results showed consistent improvement when training on small size datasets. 
\newline
While the trend of research has been bigger
and wider transformers, we show in this paper that there is still much
research to be done to make efficient networks that work on
small datasets. The presented approach of combining our modified BYOL with a ViT showed significant improvements with first and second generation, and proved to be competitive to pretraining ViTs on huge labeled datasets like JFT-300M. Our approach achieved comparable performance
while maintaining computational efficiency when working with second-generation compact transformers.
%%\paragraph{Limitations}!!!A deeper analysis with visualizations on how inputting the low-level details replaced the inability of the very first layers of the transformer encoder was left as a future work. Moreover, since the
%%goal of this paper is investigating the behaviour with extremely small size datasets, we tested our approach on STL-10 dataset only. Working with mid-range datasets like CIFAR-10 was out of the scope of this paper.
\newline
\newline
\newline
\newline
\newline
\newline
\newline

\bibliographystyle{unsrt}
\bibliography{references}
\begin{appendices}
%\appendix{Appendix}
 \renewcommand{\thetable}{\hbAppendixPrefix\arabic{table}}
 \renewcommand{\thefigure}{\hbAppendixPrefix\arabic{figure}}
 \counterwithin{table}{section}
 \counterwithin{figure}{section}
{Additional details and results from the different sections are included below.}
\section{Data augmentation of the self-supervised part}
\label{appendix:abc}
{Data augmentation is very crucial to train an effective BYOL model, for that we started by training BYOL using the same data augmentation as in the original paper \cite{grill2020bootstrap}, then we tried to modify this data augmentation in order to obtain the best possible results for the hybrid BYOL-ViT architecture. Table \ref{appendix:ab}  shows the different parameters for each data augmentation we used and Table \ref{tab:tab11} summarizes the results of training our architecture for 600 epochs. Note that one of our goals was to find the saturation point of BYOL.
\setcounter{table}{0}  
\begin{table}[htbp]
  \centering
  \caption{Data transformations used to train BYOL. Bold lines are our modifications compared to the initial Baseline used in the original paper \cite{grill2020bootstrap}}
  \label{appendix:ab}
    \begin{tabular}{|l|p{31.68em}|}
    \toprule
    Baseline &* Random Resized Crop(96)\newline{}* Random apply color jitter (brightness: 0.8,                                                       contrast: 0.8,                                                 \newline{}saturation: 0.8,                                                hue: 0.2,                                                 probability: 0.8)\newline{}* RandomHorizontalFlip()\newline{}* Random GrayScale (probability 0.2)\newline{}* Random Gaussian Blur (probability 0.2)\newline{}* Normalize(mean=[0.485,0.456,0.406],std=[0.229,0.224,0.225]) \\
    \midrule
    \midrule
    Data\_aug\_1 & * Random Resized Crop(96)\newline{}* \textbf{Random apply color jitter (brightness: 0.4,                                                 contrast: 0.4,                                                 \newline{}saturation: 0.4,                                                 hue: 0.1,                                                 probability: 0.8)}\newline{}* RandomHorizontalFlip()\newline{}* Random GrayScale (probability 0.2)\newline{}*\textbf{ Random Gaussian Blur (probability 0.5)}\newline{}*\textbf
    { Random solarizing (probability 0.2)}\newline{}*
Normalize(mean=[0.485,0.456,0.406],std=[0.229,0.224,0.225]) \\
    \midrule
    \midrule
    Data\_aug\_2 & *\textbf{ Random Rotation(15 degrees)}\newline{}* Random Resized Crop(96)\newline{}*\textbf { Random apply color jitter (brightness: 0.4,                                                 contrast: 0.4,                                              \newline{}saturation: 0.4,                                              hue: 0.1,                                            probability: 0.8)}\newline{}* RandomHorizontalFlip\newline{}* Random GrayScale (probability 0.2)\newline{}*\textbf { Random Gaussian Blur (probability 0.5)}\newline{}*\textbf{
    Random solarizing (probability 0.2)}\newline{}*
    Normalize(mean=[0.485,0.456,0.406],std=[0.229,0.224,0.225]) \\
    \midrule
    \midrule
    Data\_aug\_3 & *\textbf{ Cutout (nb=1, len=8, probability: 0.5)}\newline{}* Random Resized Crop(96)\newline{}*\textbf{ Random apply color jitter (brightness: 0.4,                                                contrast: 0.4,                                                 \newline{}saturation: 0.4,                                                 hue: 0.1,                                              probability: 0.8)}\newline{}* RandomHorizontalFlip\newline{}* Random GrayScale (probability 0.2)\newline{}*\textbf{ Random Gaussian Blur (probability 0.5)}\newline{}*\textbf{
    Random solarizing (probability 0.2)}\newline{}*
    Normalize(mean=[0.485,0.456,0.406],std=[0.229,0.224,0.225]) \\
    \midrule
    \midrule
    Data\_aug\_4 & *\textbf{ Random Rotation(15 degrees)}\newline{}*\textbf{ Cutout (nb=1, len=8, probability: 0.5)}\newline{}* Random Resized Crop(96)\newline{}*\textbf{ Random apply color jitter (brightness: 0.4,
contrast: 0.4,                                                 \newline{}saturation: 0.4,                                                hue: 0.1,                                                 probability: 0.8)}\newline{}* RandomHorizontalFlip\newline{}* Random GrayScale (probability 0.2)\newline{}*\textbf{ Random Gaussian Blur (probability 0.5)}\newline{}*\textbf{
Random solarizing (probability 0.2)}\newline{}*
Normalize(mean=[0.485,0.456,0.406],std=[0.229,0.224,0.225]) \\
    \midrule
    \midrule

    Data\_aug\_5 & * Random Resized Crop(96)\newline{}*\textbf{ Random apply color jitter (brightness: 0.4,                                                 contrast: 0.4,                                                \newline{}saturation: 0.4,hue: 0.1,                    probability: 0.8)}\newline{}* RandomHorizontalFlip\newline{}* Random GrayScale (probability 0.2)\newline{}*\textbf{ Random Gaussian Blur (probability 0.5)}\newline{}*\textbf{
    Random solarizing (probability 0.2)}\newline{}*\textbf{
    No normalization} \\
    \bottomrule
    \end{tabular}%
  \label{tab:tab22}%
\end{table}%

% Table generated by Excel2LaTeX from sheet 'Sheet2'
\begin{table}[htbp]
  \centering
  \caption{ Top-1 Accuracy in \% and Loss of our Hybrid BYOL-ViT architecture after extracting low-level features from BYOL pretrained under different data augmentations (See Table \ref{appendix:ab}) for multiple epochs. For these experiments we chose ResNet18 as BYOL's backbone.}
  \label{appendix:a}
    \begin{tabular}{|l||r|r||r|r||r|r||r|r||r|r||r|r||}
\cmidrule{2-13}    \multicolumn{1}{r||}{} & \multicolumn{2}{c||}{\textbf{Baseline}} & \multicolumn{2}{c||}{\textbf{Data aug\_1}} & \multicolumn{2}{c||}{\textbf{Data aug\_2}} & \multicolumn{2}{c||}{\textbf{Data aug\_3}} & \multicolumn{2}{c||}{\textbf{Data aug\_4}} & \multicolumn{2}{c||}{\textbf{Data aug\_5}} \\
    \midrule
    \multicolumn{1}{|p{4.775em}||}{\#epochs BYOL} & \multicolumn{1}{l|}{Acc} & \multicolumn{1}{l||}{Loss} & \multicolumn{1}{l|}{Acc} & \multicolumn{1}{l||}{Loss} & \multicolumn{1}{l|}{Acc} & \multicolumn{1}{l||}{Loss} & \multicolumn{1}{l|}{Acc} & \multicolumn{1}{l||}{Loss} & \multicolumn{1}{l|}{Acc} & \multicolumn{1}{l||}{Loss} & \multicolumn{1}{l|}{Acc} & \multicolumn{1}{l||}{Loss} \\
    \midrule
    100 epochs & 72    & 0.7755 & 74    & 0.7261 & 73.42 & 0.777 & 74.17 & 0.7223 & 74.92 & 0.7286 & 76.25 & 0.678 \\
    \midrule
    200 epochs & 74    & 0.7152 & 75.67 & 0.6467 & 74.92 & 0.7043 & 75.83 & 0.6853 & 75.33 & 0.7092 & 78.83 & 0.6 \\
    \midrule
    300 epochs & 75.1  & 0.694 & 77.17 & 0.6639 & 76.25 & 0.6811 & 76.42 & 0.6519 & 75.92 & 0.6682 & 80.33 & 0.5881 \\
    \midrule
    
    400 epochs & 75    & 0.6859 & 78.17 & 0.6469 & 77.58 & 0.654 & 76.42 & 0.6539 & 76    & 0.6945 & 79.42 & 0.5904 \\
    \midrule
    500 epochs & 74.8      &  0.684     & 77.42 & 0.6481 & 76.75 & 0.6442 & 75.92 & 0.6417 & 77.25 & 0.653 &  80.12     & 0.5894 \\
    \midrule
    600 epochs &  74.67     &  0.697     & 77.08 & 0.638 & 77.08 & 0.655 & 76.67 & 0.6463 & 75.5  & 0.6683 & 79.12      &  0.5935\\
    \midrule
    700 epochs & 74.2      &  0.699     & 77.25 & 0.6449 & 76.5  & 0.6477 & 75.58 & 0.6431 & 76.5  & 0.6446 & 79.33 & 0.601 \\
    \bottomrule
    \end{tabular}%
  \label{tab:tab11}%
\end{table}%

\section{Data augmentation of the supervised part}
\label{appendix:b}
{We did also multiple experiments to target the best data augmentation to train a Hybrid BYOL-ViT architecture, this is also crucial since we found that when it comes to the supervised part, data augmentation also matters, and should be chosen wisely as for the self-supervised part.
All the data augmentations we opted for are summarized in Table \ref{tab:app2}
\begin{table}[htbp]
  \centering
  \caption{Data augmentations used to train the hybrid BYOL-ViT architecture}
    \begin{tabular}{|c|p{20.865em}|}
    \toprule
    No Aug & \centerline{---} \\
    \midrule
    Aug\_0 & Resize((96,96)),\newline{}RandomResizedCrop((96,96),scale=(0.05,1.0)) \\
    \midrule
    Aug\_1 & Resize((96,96)),\newline{}RandomResizedCrop((96,96),scale=(0.05,1.0)),\newline{}RandomRotation(15) \\
    \midrule
    Aug\_2 & Resize((96,96)),\newline{}RandomResizedCrop((96,96),scale=(0.05,1.0)),\newline{}RandomRotation(15),\newline{}Random Gaussian Blur (probability 0.2) \\
    \midrule
    Aug\_3 & RandomCrop(96,padding=4),\newline{}RandomHorizontalFlip() \\
    \midrule
    Aug\_4 & RandomCrop(96,padding=4),\newline{}RandomHorizontalFlip(),\newline{}Random Gaussian Blur (probability 0.2) \\
    \midrule
    Aug\_5 & RandomCrop(96,padding=4),\newline{}RandomHorizontalFlip(),\newline{}Random Gaussian Blur (probability 0.2)\newline{}RandomRotation(15), \\
    \bottomrule
    \end{tabular}%
  \label{tab:app2}%
\end{table}%
\newline
We reported the results of training our architecture for 600 epochs Figure \ref{fig:plott}. Our goal was to avoid overfitting. For that, we reported training accuracy and validation accuracy, also the training loss and the validation loss. Note that for these experiments, BYOL was trained for 400 epochs with Data\_aug\_1 (Details in Table \ref{tab:tab22})}

\begin{figure}[!ht]
\centering
\includegraphics[width=1.0\textwidth]{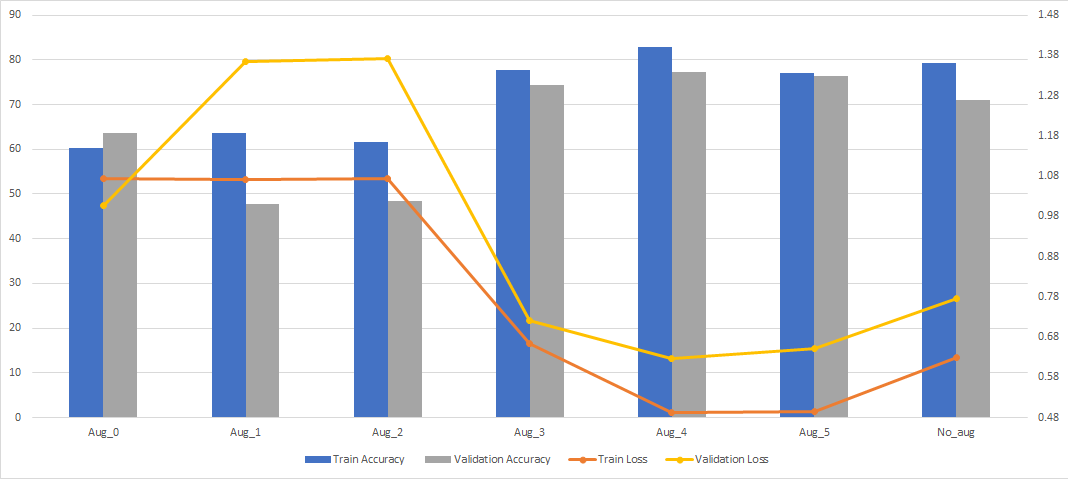}
\caption{Plots showing the impact of data augmentation in training the hybrid architecture}
\label{fig:plott}
\end{figure}

\section{Hyperparameters Tuning}
We tuned the hyperparamters per experiment and arrived at the following for each Hybrid BYOL-ViT experiment in Sec\ref{others}. We determined these hyper parameters through a parameter sweep. By picking Adam as an optimizer with a learning rate of 1e-4 and a weight decay of 0.05, using a batch size of 128
and increasing the number of epochs to 600 epochs, we achieved the best results. 
\newline

\section{LeakyReLU instead of ReLU}
\label{appendix:d}
We further study the effectiveness of replacing LeakyReLU with ReLU in the MLP architecture of BYOL. More specifically, we pretrain two BYOL models with ResNet50 as a backbone using Data\_aug\_5 (See Appendix\ref{appendix:abc}), one with ReLU layer after the BN layer, and one with LeakyReLU after the BN layer. We then extract features from Layer2 with patch size 1x1 and we feed them to a standard ViT Large. The training performances are shown in Table \ref{tab:Leaky_stl}.

% Table generated by Excel2LaTeX from sheet 'CIFAR-10'
\begin{table}[htbp]
  \centering
  \caption{Effects of ReLU and LeakyReLU in the final performance of the hybrid architecture. For these experiments, BYOL is pretrained on the set of unlabeled data provided with STL-10 dataset (100k training images).}
    \begin{tabular}{|l|r|r|r|r|}
\cmidrule{2-5}    \multicolumn{1}{r|}{} & \multicolumn{2}{c|}{\textbf{ReLU}} & \multicolumn{2}{c|}{\textbf{LeakyReLU}} \\
    \midrule
     \#Epochs to train BYOL & \multicolumn{1}{l|}{Top-1 Accuracy} & \multicolumn{1}{l|}{Loss} & \multicolumn{1}{l|}{Top-1 Accuracy} & \multicolumn{1}{l|}{Loss} \\
    \midrule
    100epochs & 75.80\%      & 0.6547      & 76.92\% \textcolor{red}{(+1.12\%)} & 0.6498 \\
    200epochs & 79.62\%      & 0.6012      & 80.33\% \textcolor{red}{(+0.71\%)} & 0.5833 \\
    300epochs & 80.31\%      & 0.5312      & 82.17\% \textcolor{red}{(+1.86\%)} & 0.5291 \\
    400epochs &  81.54\%     &  0.5075     & 83.25\% \textcolor{red}{(+1.80\%)} & 0.4979 \\
    500epochs &  81.23\%     & 0.5102      & 81.67\% \textcolor{red}{(+0.44\%)} & 0.5112 \\
    \bottomrule
    \end{tabular}%
  \label{tab:Leaky_stl}%
\end{table}%

We conduct similar experiments on small-scaled CIFAR-10 dataset \cite{site9} to test our findings in the set of unlabeled data of STL-10. We trained two BYOL models with ResNet50 as a backbone on CIFAR-10 dataset. Results are reported in Table \ref{tab:leaky_cifar}.
% Table generated by Excel2LaTeX from sheet 'CIFAR-10'
\begin{table}[H]
  \centering
  \caption{Effects of ReLU and LeakyReLU in the final performance of the hybrid architecture. For these experiments, BYOL is pretrained on CIFAR-10 (50k training images).}
    \begin{tabular}{|l|r|r|r|r|}
\cmidrule{2-5}    \multicolumn{1}{r|}{} & \multicolumn{2}{c|}{\textbf{ReLU}} & \multicolumn{2}{c|}{\textbf{LeakyReLU}} \\
    \midrule
     \#Epochs to train BYOL & \multicolumn{1}{l|}{Top-1 Accuracy} & \multicolumn{1}{l|}{Loss} & \multicolumn{1}{l|}{Top-1 Accuracy} & \multicolumn{1}{l|}{Loss} \\
    \midrule
    100epochs & 68.25\% & 0.8495 & 69.67\% \textcolor{red}{(+1.42\%)} & 0.8220 \\
    200epochs & 72.25\% & 0.7778 & 73.50\% \textcolor{red}{(+1.25\%)}  & 0.7744 \\
    300epochs & 73.08\% & 0.7462 & 72.67\% \textcolor{blue}{(-0.41\%)}      & 0.7687 \\
    400epochs & 72.58\% & 0.7674 & 74.42\% \textcolor{red}{(+1.84\%)} & 0.7505 \\
    500epochs & 72.33\% & 0.7489 &  74.42\% \textcolor{red}{(+2.09\%)}     & 0.7426 \\
    \bottomrule
    \end{tabular}
  \label{tab:leaky_cifar}
\end{table}
\section{Influence of weight decay}
\label{appendix:e}
We wanted to show the influence of changing the weight decay in the overall final performance of our hybrid BYOL-ViT architecture. Table \ref{tab:weight} summarizes some of the experiments, we started with a weight decay of 0.5 and we reduced this value by a factor of 0.1 each time. We found that reducing the weight decay slightly increases performance but greatly reduces the rate of convergence. This demonstrates that weight decay is still a useful parameter to investigate when hyperparameter tuning. We found that choosing small weight decay is helpful during early training but deteriorates final performance.
% Table generated by Excel2LaTeX from sheet 'Weight_effect'
\begin{table}[htbp]
  \centering
  \caption{Effect of weight decay. For these experiments: $LR=10^{-4}$}
    \begin{tabular}{|l|l|l|}
    \toprule
    \multicolumn{1}{|p{8em}|}{weight\_decay \newline{}coefficient} & \multicolumn{1}{p{4.045em}|}{Top-1\newline{}Accuracy} & Loss \\
    \midrule
    $5.10^{-1}$ &       \multicolumn{1}{r|}{51.25\%} & \multicolumn{1}{r|}{1.315} \\
    \midrule
    $5.10^{-2}$ & \multicolumn{1}{r|}{\textbf{ 83.25\%}} & \multicolumn{1}{r|}{\textbf{0.4979}} \\
    \midrule
    $5.10^{-3}$ &       \multicolumn{1}{r|}{82.58\%} & \multicolumn{1}{r|}{0.5384} \\
    \midrule
    $5.10^{-4}$ &       \multicolumn{1}{r|}{82.17\%} & \multicolumn{1}{r|}{0.5431} \\
    \bottomrule
    \end{tabular}%
  \label{tab:weight}%
\end{table}%
\section{Influence of Batch Size}
\label{appendix:f}
Batch size is one of the most important hyperparameters to tune in modern deep learning systems \cite{smith2018dont}\cite{hoffer2018train}. In the original paper of BYOL \cite{grill2020bootstrap}, authors wanted to use larger batch sizes to train their model as it allows computational speedups from the parallelism of GPUs and they reported better results when training with a large batch size of 4096. We wanted to test the efficiency and the effect of working with small and large batch sizes when it comes to extracting low-level details from the first layers. Results of training our Hybrid architecture with BYOL trained under different batch sizes is reported in Table \ref{tab:bs}.

% Table generated by Excel2LaTeX from sheet 'BS_effect'
\begin{table}[H]
  \centering
  \caption{Results of BYOL trained under different batch sizes, then features extracted from Layer2 with Patch 1x1 and fed to ViT\_L. Overall architecture (Hybrid BYOL-ViT) trained on STL-10 dataset}
    \begin{tabular}{|l|r|r|r|r|r|r|}
\cmidrule{2-7}    \multicolumn{1}{r|}{} & \multicolumn{2}{c|}{BS = 8} & \multicolumn{2}{c|}{BS = 16} & \multicolumn{2}{c|}{BS = 32} \\
    \midrule
    \#Epochs & \multicolumn{1}{l|}{Top-1 Accuracy} & \multicolumn{1}{l|}{Loss} & \multicolumn{1}{l|}{Top-1 Accuracy} & \multicolumn{1}{l|}{Loss} & \multicolumn{1}{l|}{Top-1 Accuracy} & \multicolumn{1}{l|}{Loss} \\
    \midrule
    100ep &  74.92\%     & 0.6946      & 76.92\% & 0.6498 & 77.67\%\textcolor{red}{(+0.75)} & 0.6042 \\
    200ep & 79.83\%      & 0.5765      & 80.33\% & 0.5833 & 80.17\%\textcolor{blue}{(-0.16)} & 0.5551 \\
    300ep &  81\%     &   0.5183    & 82.17\% & 0.5291 & 81.83\%\textcolor{blue}{(-0.34)} & 0.5144 \\
    400ep &  83.25\%     &  0.5095     & 83.25\% & 0.4979 & 80.92\%\textcolor{blue}{(-2.33)} & 0.5371 \\
    500ep & 81.25\%      & 0.5098      & 81.67\% & 0.5112 & 79.92\%\textcolor{blue}{(-1.75)} & 0.5536 \\
    \midrule
    \multicolumn{1}{r}{} & \multicolumn{1}{r}{} & \multicolumn{1}{r}{} & \multicolumn{1}{r}{} & \multicolumn{1}{r}{} & \multicolumn{1}{r}{} & \multicolumn{1}{r}{} \\
\cmidrule{2-7}    \multicolumn{1}{r|}{} & \multicolumn{2}{c|}{BS = 64} & \multicolumn{2}{c|}{BS = 128} & \multicolumn{2}{c|}{BS = 256} \\
    \midrule
    \#Epochs & \multicolumn{1}{l|}{Top-1 Accuracy} & \multicolumn{1}{l|}{Loss} & \multicolumn{1}{l|}{Top-1 Accuracy} & \multicolumn{1}{l|}{Loss} & \multicolumn{1}{l|}{Top-1 Accuracy} & \multicolumn{1}{l|}{Loss} \\
    \midrule
    100ep & 77.17\%\textcolor{red}{(+0.25)} & 0.6612 & 78.58\%\textcolor{red}{(+1.66)} & 0.6295 & 74\%\textcolor{blue}{(-2.92)} & 0.7279 \\
    200ep & 79.58\%\textcolor{blue}{(-0.75)} & 0.5915 & 80.08\%\textcolor{blue}{(-0.25)} & 0.5916 & 78.83\%\textcolor{blue}{(-1.50)} & 0.6556 \\
    300ep & 79.25\%\textcolor{blue}{(-2.92)} & 0.5725 & 79.25\%\textcolor{blue}{(-2.92)} & 0.5966 & 77.83\%\textcolor{blue}{(-4.34)} & 0.6357 \\
    400ep & 77.75\%\textcolor{blue}{(-5.50)} & 0.6044 & 78.33\%\textcolor{blue}{(-4.92)} & 0.6162 & 76.58\%\textcolor{blue}{(-6.67)} & 0.6556 \\
    500ep & 77.92\%\textcolor{blue}{(-3.75)} & 0.6113 & 78.67\%\textcolor{blue}{(-3.00)} & 0.6019 & 77.58\%\textcolor{blue}{(-4.09)} & 0.6473 \\
    \bottomrule
    \end{tabular}
  \label{tab:bs}
\end{table}}
\end{appendices}

\end{document}